\documentclass[conference, 9pt]{IEEEtran}

\usepackage{subcaption}
\usepackage{graphicx}
\graphicspath{{./figs/}}
\usepackage{amsmath}
\usepackage{booktabs}
\usepackage{multirow}
\usepackage{siunitx}
\usepackage{xspace}

\usepackage{algorithm}
\usepackage{algpseudocode}
\usepackage[dvipsnames]{xcolor}
\usepackage[colorlinks]{hyperref}
\usepackage{stfloats}
\usepackage[font=small]{caption}

\hypersetup{linkcolor=blue,
,citecolor=magenta,
,filecolor=Mulberry
,urlcolor=NavyBlue
}

\usepackage[square,sort,comma,numbers,compress]{natbib}

\begin{document}

\newcommand{\HZ}[1]{\textcolor{blue}{[HZ: #1]}}
\newcommand{\DrYoOh}[1]{\textcolor{blue}{[DrYoOh: #1]}}

\renewcommand{\bibfont}{\small} 
\newcommand{\newhuge}{\fontsize{23}{24}\selectfont}

\newtheorem{remark}{Remark}

\title{\vskip -0.2cm HeaRT: A \underline{H}i\underline{e}r\underline{a}rchical Circuit \underline{R}easoning \underline{T}ree-Based Agentic Framework for AMS Design Optimization
\vspace{-0.2cm}}

\author{Souradip Poddar\textsuperscript{1}, 
Chia-Tung Ho\textsuperscript{2}, 
Ziming Wei\textsuperscript{1}, 
Weidong Cao\textsuperscript{3},
Haoxing Ren\textsuperscript{2}, 
David Z. Pan\textsuperscript{1}\\
\textsuperscript{1}ECE Department, The University of Texas at Austin \quad 
\textsuperscript{2}NVIDIA Corporation \quad
\textsuperscript{3}The George Washington University\\
\{souradippddr1@,
dpan@ece.\}utexas.edu
\vspace{-0.2cm}
}

\maketitle
\begin{abstract}
Conventional AI-driven AMS design automation algorithms remain constrained by their reliance on high-quality datasets to capture underlying circuit behavior, coupled with poor transferability across architectures, and a lack of adaptive mechanisms.
This work proposes HeaRT, a hierarchical circuit reasoning-based agentic framework for automation loops and a step toward adaptive, human-style design optimization. 
HeaRT consistently improves $F_{1,\mathrm{subcircuits}}$ by $\geq13.5\%$ and $F_{1,\mathrm{loops}}$ by $\geq37.8\%$ over few-shot prompting baselines across multiple LLM backbones on our 40-circuit AMS benchmark of flattened SPICE netlists, 
even as circuit complexity increases. 
Our experiments further show that HeaRT achieves $\geq3\times$ faster convergence in incremental design adaptation tasks under specification shifts across diverse optimization approaches, supporting both topology reconfiguration and sizing.
\end{abstract}
\section{Introduction} \label{sec:Introduction}
\sloppy
Analog and mixed-signal (AMS) circuit design remains challenging to automate due to its fully custom flows, intricate trade-offs in deep sub-micron technologies, and the high cost of re-optimization when specifications change.
Conventional Bayesian Optimization (BO) methods ~\cite{lyu2018batch,BO_sizing_Shuhan,wang2023recent} offer strong sample efficiency but struggle to scale effectively in complex, high-dimensional design spaces.
Recent learning-based approaches, particularly reinforcement learning (RL) ~\cite{DATE20:Settaluri:Autockt, DAC20:Wang:GCNRL, li2021circuit,
DAC21:Budak:DNN_Opt, ASPDAC23:Budak:APOSTLE, choi2023reinforcement, zhang2023automated,  DATE24:CRONus:Oh, sun2024evdmarl, bao2024multiagent}, demonstrate improved scalability for larger circuits, 
yet incur prohibitive simulation costs.
Moreover, their reliance on manually defined functional decompositions or handcrafted sub-block reward formulations limits autonomy and scalability ~\cite{sun2024evdmarl, bao2024multiagent}.
Being purely data-driven, these models exhibit an inherent \textit{black-box} nature~\cite{gielen2025invited}, hindering their ability to capture circuit intuition or physical causality.
Consequently, they struggle to adapt to incremental architectural changes, often requiring new data collection, 
and the lack of explainability of the design trade-offs considered erodes designer confidence in the quality of the results ~\cite{gielen2025brief} (Fig.~\ref{fig:HeaRT_Motivation}(a)).

Most recently, Large Language Model (LLM)-based methods~\cite{chen2024artisan,yin2024ado,lai2025analogcoder,lai2026analogcoderpro,atelier_tcad_2025,zhang2025analogxpert,gao2025analoggenie,gao2025analoggenielite,kochar2025ledro,xu2025image2net, chen2025analogtester,gielen2025invited} have shown promise in advancing AMS design automation.
By leveraging reasoning and agentic capabilities, these models can emulate key aspects of human design workflows.
However, current vanilla LLM-based approaches fail to incorporate the 
hierarchical cognitive structure underlying expert AMS design ~\cite{razavi2000design,kundert2003principles,gielen2002modeling},
resulting in unstructured and inconsistent reasoning that limits both their credibility and practical effectiveness (Fig. ~\ref{fig:HeaRT_Motivation}(b)).
Moreover, existing automation methods lack explicit mechanisms to adaptively balance design reuse and redesign within topology-sizing co-optimization, thereby treating each specification update as a new problem.
In turn, they re-optimize entire circuits from scratch, 
leading to {\textit{catastrophic forgetting}}~\cite{kirkpatrick2017overcoming} of previously acquired design knowledge.
In real AMS workflows, many subcircuits are already layout-planned, variation-optimized~\cite{budak2023joint,budak2023practical,shi2022robustanalog,gao2023rose,cao2024rose}, or even silicon-proven, making such full re-optimization impractical.
This lack of architectural and contextual awareness causes redundant computation, degraded sample efficiency, and inconsistent reliability, limiting the practical deployability of current automation frameworks in industrial design flows.

To fully harness the potential of LLMs 
for AMS automation, we propose HeaRT, 
a multi-level agentic reasoning framework
that enables reasoning-grounded downstream applications.
HeaRT draws inspiration from the hierarchical abstraction principles inherent to human circuit design, constructing a hierarchical circuit reasoning tree to enable
structured, context-aware reasoning
with query-conditioned traversal paths for improved interpretability and debugging~\cite{liu2025ic+} (Fig.~\ref{fig:HeaRT_Motivation}(c)).
Our key contributions are summarized as follows:

 \begin{figure}[t!]
    \centering
    \includegraphics[width=0.95\columnwidth]{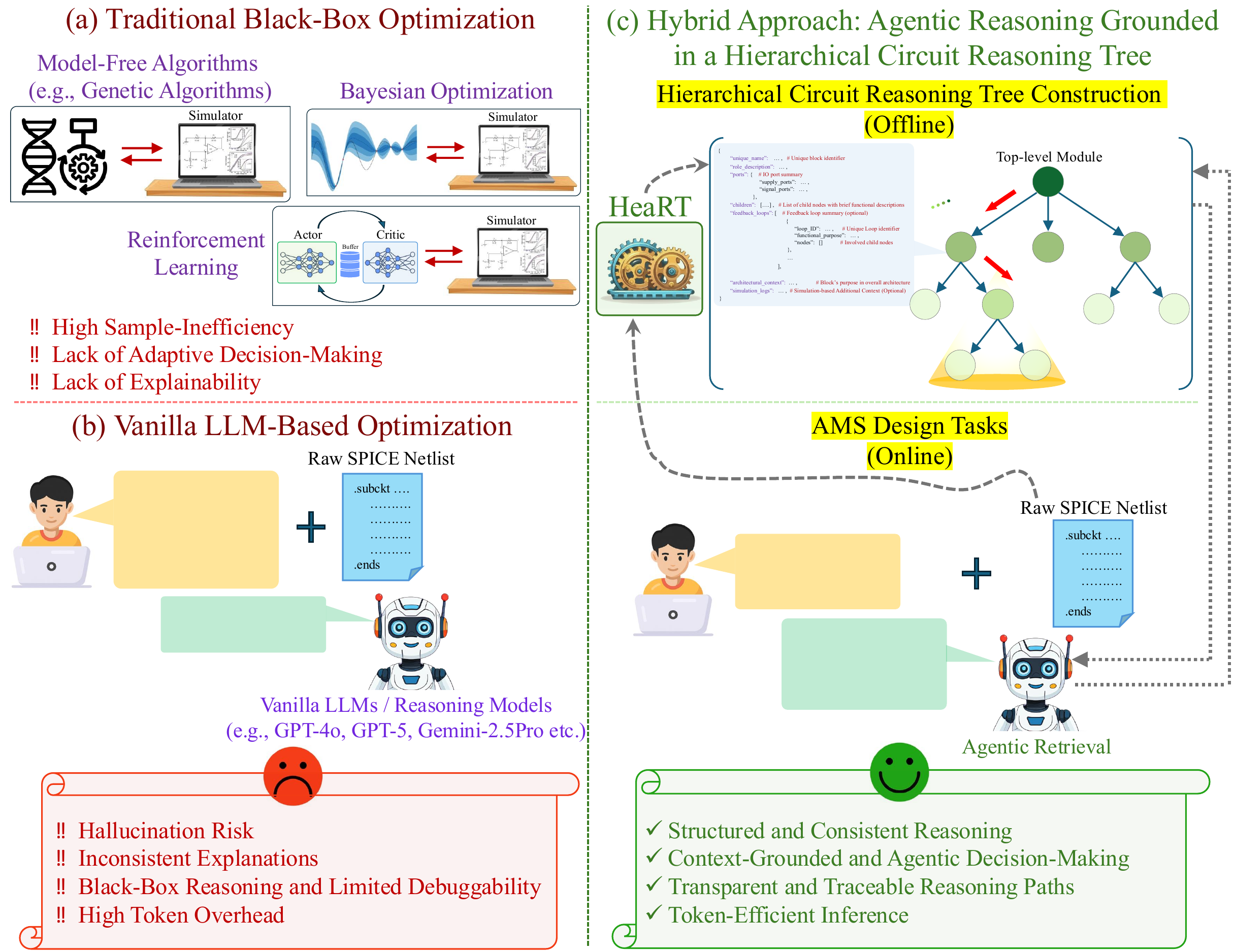}
    \vspace{-2pt}
    \caption{An illustration of comparisons of traditional black-box optimization algorithm, vanilla LLM-based approach, and the proposed HeaRT framework for the AMS design task.}
    \vspace{-14pt}
    \label{fig:HeaRT_Motivation}
\end{figure}

\begin{figure*}[!t]
    \centering
    \includegraphics[width=\linewidth]{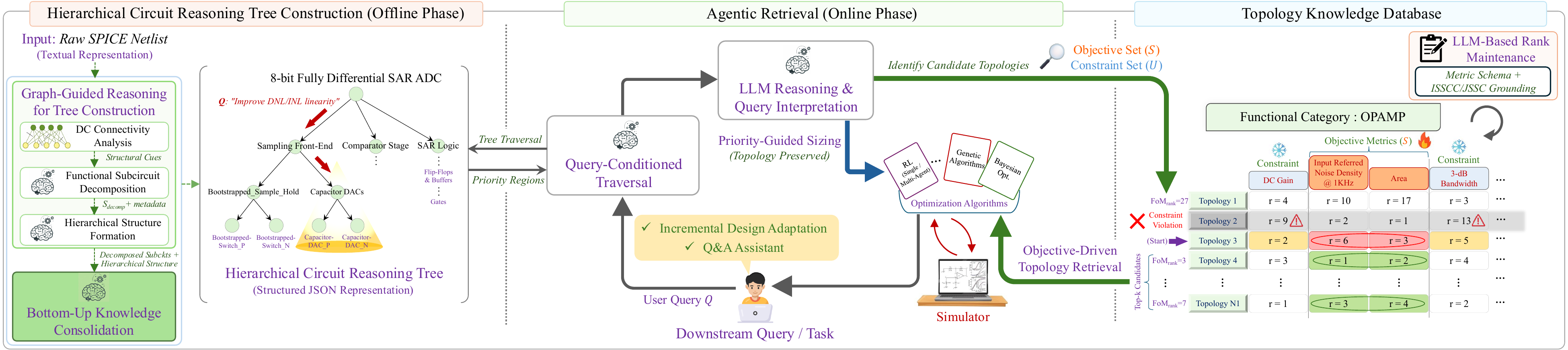}
    \vspace{-14pt}
    \caption{Overall workflow of the HeaRT framework.
    Left: offline hierarchical circuit reasoning tree construction from SPICE netlists via top-down graph-guided subcircuit identification, hierarchical abstraction, and bottom-up knowledge consolidation. 
    Middle: query-conditioned traversal and agentic retrieval for AMS design tasks.
    Right: topology knowledge database with metric-wise ranking for objective-driven topology retrieval.
    }
    \label{fig:algorithm_steps}
    \vspace{-14pt}
\end{figure*}

\begin{itemize}
\item 
We develop HeaRT, a multi-level agentic reasoning framework that performs 
graph-guided top-down circuit decomposition with hierarchical organization, followed by bottom-up knowledge consolidation into a hierarchical 
circuit reasoning tree, 
enabling structured, context-aware LLM reasoning for AMS design tasks.

\item We construct a 40-circuit AMS benchmark repository spanning multiple circuit families and complexity tiers, together with expert-annotated ground-truth labels. 
The benchmark will be publicly released upon acceptance.

\item We evaluate HeaRT across multiple LLM backbones on this benchmark and show that it consistently boosts $F_{1,\mathrm{subcircuits}}$ by $\geq$13.5\% and $F_{1,\mathrm{loops}}$ by $\geq$37.8\% over few-shot prompting baselines. 
When integrated with existing optimization algorithms, HeaRT further enables $\geq3\times$ faster convergence in incremental design-adaptation tasks involving both sizing and topology reconfiguration.

\end{itemize}

\section{Preliminaries} \label{sec:prelim}
\subsection{LLM-Aided AMS Design Automation}\label{subsec:reasoning_models}

The emergence of reasoning-enabled LLMs endowed with multimodal understanding and agentic decision-making capabilities ~\cite{wei2022CoT, wang2024survey, jiang2024llmopt, xi2025rise, wu2025agentic,xu2025towards,chen2025towards, guo2025deepseek, schmid2025new} has redefined autonomous problem-solving across diverse domains, motivating their evaluation within EDA.
Critically, expert AMS designers reason about unseen circuit structures through hierarchical cognitive processes of abstraction, analysis, and bottom-up interpretation ~\cite{razavi2000design,kundert2003principles,gielen2002modeling}. 
However, existing LLM-based AMS approaches do not explicitly exploit this structured reasoning paradigm, leading to inconsistencies, hallucinations~\cite{huang2025hallucinationsurvey}, and limited reasoning traceability that conflict with the precision, determinism, and verifiability required in circuit design ~\cite{liu2025ic+}.
Moreover, recent LLM-aided efforts ~\cite{chen2024artisan,yin2024ado,lai2025analogcoder,atelier_tcad_2025,zhang2025analogxpert,gao2025analoggenie,gao2025analoggenielite,kochar2025ledro,xu2025image2net, chen2025analogtester,gielen2025invited} remain task-specific, 
often deploying LLMs within narrowly scoped optimization roles rather than leveraging their broader reasoning capabilities.
EEsizer~\cite{liu2025eesizer} employs Chain-of-Thought (CoT) reasoning~\cite{wei2022CoT} with simulation-in-the-loop, 
operating primarily at the parameter level for sizing rather than structural circuit interpretation.
Similarly,
~\cite{ns2025ai} uses 
metric-driven identification of local transistor patterns (e.g., diff pairs, current mirrors), 
in contrast to the topology-centric hierarchical abstraction used by human designers,
where structural reasoning precedes consideration of performance metrics.
These limitations highlight the need for structured reasoning frameworks that align LLM capabilities with human design reasoning.

\subsection{Structural Interpretation of AMS Circuits}\label{subsec:need_decomp}

Flattened SPICE netlists represent circuits as a single connectivity graph where devices share arbitrary interconnections, 
obscuring the structural organization designers rely on to interpret circuit functionality. 
Structural abstraction is therefore necessary for modular analysis and optimization, thereby better unlocking the capabilities of LLM in AMS design. 
Existing subcircuit identification methods rely on hand-crafted rule-based ~\cite{abel2022functional}, template library-based ~\cite{li2024efficient}, learning-based ~\cite{kunal2020gana,kunal2023gnn}, or class-specific few-shot   approaches~\cite{pham2025genie}. 
These methods remain constrained by template coverage or training data distribution and primarily focus on identifying local transistor patterns.
In this paper, we use the term \textit{subcircuits} to denote modular functional blocks corresponding to the block-level abstractions commonly used by analog designers. 
Examples include gain stages in amplifiers; 
resistor dividers, error amplifiers, and pass devices in LDO regulators~\cite{razavi2000design}.
However, given the diversity of analog circuit architectures, automated structural partitioning
to recover high-level functional abstractions remains challenging~\cite{pham2025genie},
since electrically coupled fragments hinder independent functional interpretation and disrupt feedback loop relationships.
When correctly identified, such abstractions preserve subcircuit-level functional integrity and expose the modular structure required for hierarchical circuit reasoning and systematic interpretation of circuit blocks and their feedback interactions~\cite{gielen2002modeling,kundert2003principles}.
To quantitatively evaluate the correctness of subcircuit decomposition and feedback loop identification, we introduce two F1-score-based metrics:

\begin{equation}
F_{1,\mathrm{subcircuits}} = 
\frac{2|\mathcal{S} \cap \mathcal{S}^*|}
{|\mathcal{S}| + |\mathcal{S}^*|}, 
\qquad
F_{1,\mathrm{loops}} = 
\frac{2|\mathcal{L} \cap \mathcal{L}^*|}
{|\mathcal{L}| + |\mathcal{L}^*|},
\label{eq:f1_scores}
\end{equation}

\noindent
where $\mathcal{S}^*$ and $\mathcal{S}$ denote the ground-truth and identified subcircuit sets, respectively, and $\mathcal{L}^*$ and $\mathcal{L}$ denote the ground-truth and identified feedback loop sets.
A subcircuit match requires agreement in device membership and internal net connectivity, 
whereas a feedback loop match requires agreement in participating devices (or hierarchical blocks), loop polarity, and functional role.

\subsection{Performance Figure of Merit (FoM)}

Given $m$ normalized performance metrics
$\{ f_r(x) \}_{r=1}^{m}$ defined over their acceptable bound intervals 
$S_r^{\mathrm{bound}} = [L_r, H_r]$,
with $f_r(x) \ge 0$ indicating satisfaction,
we define the performance Figure of Merit (FoM) as follows:

\begin{equation}
\mathrm{FoM}(x)
=
\sum_{r=1}^{m}
w_r \,
\min\!\left(1,\max\!\left(0, f_r(x)\right)\right),
\label{eq:perf_fom}
\end{equation}
where $w_r$ is the weighting factor.
The $\max(\cdot)$ operator truncates negative values arising from
constraint violation, while the $\min(\cdot)$ operator bounds
each metric contribution, preventing any single term from
dominating the FoM beyond full satisfaction.

\subsection{Design Effort Reuse Under Specification Shift}\label{subsec:eval_metrics_CEREBRO}
To complement the performance-centric FoM, 
it is also essential to assess how effectively an optimization process preserves prior design effort
in scenarios where incremental design adaptation is preferred over full re-optimization.
In practical AMS workflows, re-optimizing solely for performance under updated specifications, without contextual awareness, 
may inadvertently discard valuable prior efforts, such as robustness across PVT corners, device mismatch, and layout parasitics.
Hence, analogous to parameter drift induced by fine-tuning in machine learning literature ~\cite{kirkpatrick2017overcoming, kim2023achieving, zenke2017continual}, 
we introduce a design effort reuse score, denoted by $\mathcal{R}_{\mathrm{effort}}$, to quantify deviation of design variables under specification shifts within a fixed architecture relative to a previously validated reference design, defined as:

\begin{equation}
\mathcal{R}_{\mathrm{effort}} = {1}/{M} \sum_{j=1}^{M}
\Big[1 - \min\!\left(1,\left|\log_{10}(x_j/x_j^{(0)})\right|\right)\Big].
\label{eq:trs_dvrs}
\end{equation}

\noindent
Here, $M$ is the number of design variables,
and $x_j^{(0)}$ and $x_j$ represent the initial and modified values of the $j$-th design variable, respectively.
The $\min(\cdot,1)$ cap prevents single large deviations from dominating the score.
$\mathcal{R}_{\mathrm{effort}} \in [0, 1]$, where $1$ indicates complete parameter reuse within the fixed architecture and 
smaller values
reflect substantial deviation from the validated reference design.

\section{The HeaRT Framework} \label{sec:algo}

Current LLM-based approaches for AMS design fall short of the explainability and structured reasoning exhibited by human design flows, and their opaque, inconsistent reasoning limits trustworthiness and practical deployment.
Inspired by the cognitive process through which human designers analyze, abstract, and interpret circuits, we propose HeaRT as a hierarchical reasoning framework that leverages human-inspired design philosophy to achieve transparent, context-consistent circuit understanding for downstream agentic tasks.
As depicted in Fig.~\ref{fig:algorithm_steps} (left),
HeaRT begins with 
graph-guided LLM-assisted subcircuit extraction, followed by hierarchical organization and bottom-up architectural knowledge consolidation. 
By combining top-down circuit decomposition with bottom-up reasoning, the framework mirrors the hierarchical abstraction principles used in circuit design.
Local metadata provides informative cues, while access to the full netlist as global context maintains semantically consistent multi-level circuit reasoning.
The framework of HeaRT operates through two complementary phases: an offline, one-time knowledge-building stage that constructs the hierarchical circuit reasoning tree (Section~\ref{HeaRT_Phase_1}), and an online, real-time agentic retrieval stage that leverages standard LLM reasoning capabilities while grounding responses in this tree,
enabling structured,
context-grounded inference beyond single-shot full-circuit reasoning
(Section ~\ref{HeaRT_Phase_2}).
The following subsections describe each phase in further detail.
To promote reproducibility and further research, we will release HeaRT as an open-source framework upon acceptance.

\begin{figure}[t]
    \centering
    \includegraphics[width=0.95\linewidth]{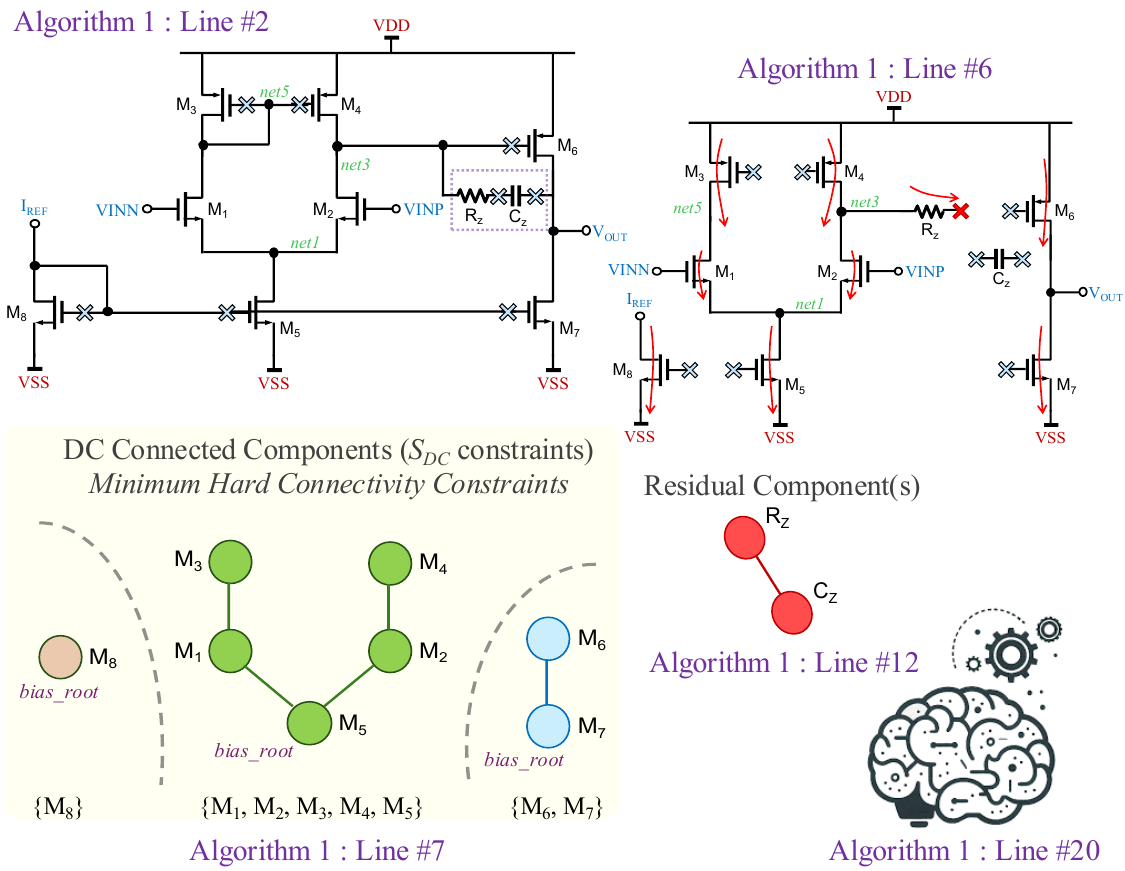}
    \caption{\small An illustrative example of HeaRT's circuit decomposition.}
    \label{fig:algorithm_phases_overview}
    \vspace{-14pt}
\end{figure}

\subsection{Hierarchical Circuit Reasoning Tree Construction}\label{HeaRT_Phase_1}
This phase constitutes a one-time offline process that constructs a hierarchical circuit reasoning tree directly from raw SPICE netlists, 
grounding subsequent agentic retrieval inference tasks.

\subsubsection{Graph-Guided Reasoning for Tree Construction}

We first transform the SPICE-level circuit netlist into a device–net bipartite graph 
$G = (V_D, V_N, E)$, where device nodes $V_D$ represents circuit components (e.g., transistors, resistors, capacitors), and net nodes $V_N$ correspond to unique net names.
Each net node is labeled by its type as \texttt{SUPPLY\_PORT}, \texttt{SIGNAL\_PORT}, or \texttt{INTERNAL\_NET}, and
each edge $e \in E$ encodes a unique device-terminal-net connection. 
For decomposition purposes, MOSFETs are simplified to three-terminal devices (\texttt{D}, \texttt{G}, \texttt{S}), omitting body connections.
From this graph, we extract DC-connected components that capture conductive paths between supply rails, 
exposing the circuit’s biasing structure commonly used by designers to reason about functional organization. 
These components serve as structural cues for LLM-assisted circuit decomposition
into the functional subcircuits defined
in Section~\ref{subsec:need_decomp}, with metadata generated via Chain-of-Thought (CoT) reasoning~\cite{wei2022CoT} and in-context examples~\cite{brown2020language} as described in Algorithm~\ref{alg:splitter}.
Fig.~\ref{fig:algorithm_phases_overview} illustrates an example of this decomposition.
The extracted subcircuits, $\mathbf{S_{decomp}}$, together with their metadata 
(e.g., $\texttt{subcircuit\_ID}$, $\texttt{role\_hint}$, architectural context, port annotations, etc.),
are provided to a second LLM stage to construct the hierarchical reasoning tree for subsequent bottom-up knowledge consolidation.

\subsubsection{Bottom-Up Knowledge Consolidation}

In this step, the LLM performs bottom-up hierarchical reasoning from leaf to root,
aggregating subcircuit-level reasoning and propagating structural and functional insights upward through the hierarchy.
Local and global metadata guide 
contextual consistency, 
enabling the model to interpret each subcircuit’s role relative to its parent and the broader circuit. 
At each node, the LLM identifies local feedback loops among its immediate children and assigns unique loop IDs where applicable.
As reasoning progresses upward, annotated netlists with corresponding signal and supply ports are synthesized and stored at each node, 
forming consistent
representations across abstraction levels.
Finally, all inferred functional relations are consolidated into the hierarchical circuit reasoning tree for subsequent retrieval during inference.

\vspace{-3pt}
\begin{algorithm}[h]
\scriptsize
\caption{\small LLM-Assisted Functional Subcircuit Decomposition}
\label{alg:splitter}
\begin{algorithmic}[1]

\Require Raw SPICE netlist $\mathcal{N}$
\Ensure Subcircuits $\mathcal{S}_{\text{decomp}}$ enriched with contextual metadata

\vspace{3pt}
\Function{\textcolor{blue}{ExtractDCSubcircuits}}{$G$}
    \State $G_{\text{DC}} \gets$ remove all non–DC-conductive terminals (MOS gates, capacitor nodes), then drop isolated nodes

    \State Initialize $\mathcal{R}_{\mathrm{VDD}} \gets \emptyset$, 
       $\mathcal{R}_{\mathrm{GND}} \gets \emptyset$, 
       $D_{\mathrm{DC Alive}} \gets \emptyset$

    \State Perform multi-source BFS on $G_{\mathrm{DC}}$ from \{\texttt{VDD}, \texttt{GND}\}
    \State $\mathcal{R}_{\mathrm{VDD}},\,\mathcal{R}_{\mathrm{GND}} \gets$ terminals reachable from \texttt{VDD} and \texttt{GND}, respectively

    \State $D_{\mathrm{DC\,alive}} \gets 
\{\, d=(n_1,n_2)\in D \mid 
      (n_1\in \mathcal{R}_{\mathrm{VDD}} \land n_2\in \mathcal{R}_{\mathrm{GND}}) 
      \;\lor\;
      (n_2\in \mathcal{R}_{\mathrm{VDD}} \land n_1\in \mathcal{R}_{\mathrm{GND}})
   \,\}$

    \State $\mathcal{S}_{DC} \gets$ connected components formed by DC-alive devices
    \State \Return $\mathcal{S}_{DC}$
\EndFunction

\vspace{3pt}
\Function{\textcolor{blue}{ProcessResidualAndACComponents}}{$G, \mathcal{S}_{DC}$}
    \State $G_{\mathrm{res}} \gets G \setminus \mathcal{S}_{DC}$
    \State $\mathcal{S}_{AC} \gets$ connected components of $G_{\mathrm{res}}$
    \State \Return $\mathcal{S}_{AC}$
\EndFunction

\vspace{3pt}
\State $G(V_D, V_N, E) \gets \textcolor{blue}{\textsc{NetlistToBipartite}}(\mathcal{N})$
      \Comment{device--net bipartite graph construction}
\State Annotate each $n \in V_N$ as 
       \texttt{SUPPLY\_RAIL}, \texttt{SIGNAL\_PORT}, or \texttt{INTERNAL\_NET}
       \Comment{initial net-level metadata}
\State $\mathcal{S}_{DC} \gets 
       \textcolor{blue}{\textsc{ExtractDCSubcircuits}}(G)$
       \Comment{DC Connectivity Cues}
\State $\mathcal{S}_{AC} \gets 
       \textcolor{blue}{\textsc{ProcessResidualAndACComponents}}(G,\,\mathcal{S}_{DC})$
\State $\mathcal{P} \gets \textcolor{blue}{\textsc{BuildPrompt}}(\mathcal{N},\,\mathcal{S}_{DC},\,\mathcal{S}_{AC},\,\text{CoT},\,\text{In-Context Exemplars})$

\State $O_{\mathrm{LLM}} \gets \textcolor{blue}{\textsc{LLM}}(\mathcal{P})$
\Comment{LLM Reasoning for Circuit Decomposition}
\State $\mathcal{S}_{\mathrm{decomp}} \gets \mathrm{Parse}(O_{\mathrm{LLM}})$

\State \Return $\mathcal{S}_{\mathrm{decomp}}$
\end{algorithmic}

\end{algorithm}
\vspace{-10pt}

\subsection{Agentic Retrieval}
\label{HeaRT_Phase_2}
This phase performs online, real-time agentic traversal and retrieval, grounding LLM reasoning in the hierarchical circuit reasoning tree built during the offline setup.
The retrieval operations are bounded by the tree depth, 
reducing exploration overhead and supporting consistent, context-grounded inference.
To reduce token usage while maintaining reasoning accuracy,
we employ context engineering strategies ~\cite{schmid2025new}, including context compression,
local–global context coordination,
and selective context reuse.

\subsubsection{Topology Knowledge Database Management}
\label{knowledge_database_management_SP}
As illustrated in Fig. ~\ref{fig:algorithm_steps} (right),
we maintain a topology knowledge database in which circuits are organized by functional category.
Each row corresponds to a circuit topology $\mathcal{T}$ represented by its normalized circuit graph,
while columns correspond to the family-specific performance metrics defined in our circuit-family metric schema (e.g., 3-dB bandwidth, DC gain, and phase margin for OPAMPs; temperature coefficient and startup margin for bandgap references, and so on). 
Each topology entry stores a metric-wise rank $r_i(\mathcal{T})$ that reflects its relative performance capability along metric $i$.
The database is initialized and updated using LLM reasoning grounded in analog design literature, including representative circuits and design insights from ISSCC and JSSC publications, as well as standard textbooks ~\cite{razavi2000design,gray2024analysis}, 
in conjunction with the circuit-family metric schema.
For each topology, the LLM assigns metric-wise rankings along with concise justification cues reflecting expected performance trade-offs.
During updates, the LLM conditions on the current repository context to provide additional grounding and ensure consistency across entries.
New topologies are inserted into existing metric-wise ranking lists via ordered-list insertion, while newly introduced performance metrics are added as new ranking columns and populated accordingly.
New circuit families follow the same ranking procedure under their respective metric schemas.
To support fast lookup,
each topology entry additionally stores a Weisfeiler–Lehman (WL) graph fingerprint ~\cite{shervashidze2011weisfeiler} as an auxiliary hash index.
This database 
supports task-specific rank-based retrieval during downstream applications.

\begin{table*}[t]
\centering
\small
\caption{Comparison of LLM-Based Methods for Analog and Mixed-Signal (AMS) Design.}
\vspace{-4pt}
\renewcommand{\arraystretch}{0.75}
\setlength{\tabcolsep}{13pt}
\scriptsize
\begin{tabular}{lccccc}
\toprule
\multirow{2}{*}{\textbf{Framework}} &
\multicolumn{1}{c}{\textbf{Multiple}} &
\multicolumn{1}{c}{\textbf{Hierarchical}} &
\multicolumn{1}{c}{\textbf{Performance Objective-}} &
\multicolumn{1}{c}{\textbf{Targeted Opt.}} &
\multicolumn{1}{c}{\textbf{Commercial}}
\\
 &  \textbf{Types$^{1}$} & \textbf{Circuit Reasoning} &  \textbf{Driven Topology Opt.$^{2}$}  &  \textbf{Tasks} & \textbf{PDKs$^{3}$}\\
\midrule

Artisan ~\cite{chen2024artisan}                  & -- & --    & $\bullet$ & TR \& Sizing & -- \\

ADO-LLM ~\cite{yin2024ado}               & $\bullet$ & --  & --  & Sizing Only &  $\bullet$ \\

AnalogCoder ~\cite{lai2025analogcoder}             & $\bullet$ & --    & --   & TG & -- \\

Atelier ~\cite{atelier_tcad_2025}                  & -- & -- & $\circ$ & TG$^\dagger$ \& Sizing & $\bullet$ \\

AnalogXPert ~\cite{zhang2025analogxpert}           & -- & --   & -- & TG$^\ddagger$ & -- \\

AnalogGenie/Lite ~\cite{gao2025analoggenie,gao2025analoggenielite}  & $\bullet$  & -- & -- & TG \& Sizing & -- \\

LEDRO ~\cite{kochar2025ledro}                 & -- & --  & --  & Sizing Only &  -- \\

MenTeR ~\cite{chen2025menter}                 & $\bullet$ & --  & $\bullet$  & TG \& Sizing &  --

\\

ADO-KT ~\cite{ns2025ai}                 & $\bullet$ & --  & --  & Sizing Only &  --
\\

AnaFlow ~\cite{gielen2025invited}       & -- & --  & --  & Sizing Only & --\\

AnalogCoder-Pro ~\cite{lai2026analogcoderpro}       & $\bullet$  & -- & $\circ$  & TG \& Sizing & --\\

\textbf{HeaRT (This Work)}        & $\bullet$ & $\bullet$  & $\bullet$  & TR \& Context-Aware Sizing & $\bullet$ \\
\bottomrule
\end{tabular}
\vspace{-4pt}
\begin{flushleft}
\scriptsize
$^{1}$~Demonstrates support across multiple AMS circuit families, 
$^{2}$~Supports target performance-conditioned topology optimization, 
$^{3}$~ Demonstrates evaluation using commercial foundry PDKs,
$^\dagger$~Topology generation via retrieval followed by template-based mutation rather than synthesizing from scratch, 
$^{\ddagger}$~Retrieved subcircuits from circuit library combined to generate larger composite architectures,
\textbf{TR}: Topology Retrieval/Selection, 
\textbf{TG}: Topology Generation,
$\bullet$~Supported, $\circ$~Partial,  --~Not Supported.
\end{flushleft}
\label{tab:llm_comparison}
\vspace{-18pt}
\end{table*}

\subsubsection{Reasoning-Guided and Query-Conditioned Traversal}
In this stage, the agent performs a query-conditioned traversal of the hierarchical circuit reasoning tree.
A single LLM pass first assigns query-conditioned relevance weights to every edge using CoT and in-context reasoning.
Before traversal, each node $v$ is evaluated using the following branch-admission criterion, which determines whether its children should be enqueued during BFS:
\begin{equation}
\small
\begin{aligned}
\text{Branch\_Cut}(v) =\;&
\left\{
\begin{aligned}
&\max_{u \in \mathrm{Children}(v)} w_{(v,u)} < \tau_{\mathrm{stop}}
\\[4pt]
&\text{or}\;
\max_{u \in \mathrm{Children}(v)} w_{(v,u)} - \min_{u \in \mathrm{Children}(v)}
<
\epsilon
\end{aligned}
\right\},
\end{aligned}
\label{eq:stop_condition}
\end{equation}
\noindent
where $w_{(v,u)}$ is the query-conditioned relevance of child $u$, 
$ \tau_{\mathrm{stop}}$ halts expansion when all children are weak, and 
$\epsilon$ stops expansion when all children are similarly strong (i.e., no dominant direction).
If condition ~\ref{eq:stop_condition} is satisfied at a node $v$,
all of its children are marked as non-admissible and therefore never enqueued into the BFS queue. 
A BFS from the root then enqueues children only when the Branch\_Cut criterion is not met.
Traversal thus naturally terminates either at suppressed nodes or true leaves.
The resulting root-to-terminal traversal paths define query-conditioned priority search regions, guiding the framework toward the most influential design variables for downstream tasks such as sizing optimization or objective-driven retrieval while maintaining full circuit context.

\subsubsection{Objective-Driven Retrieval for Topology Optimization}
Given a topology optimization task, HeaRT performs query-conditioned BFS traversal
to identify candidate subcircuits for objective-driven retrieval. 
For each identified subcircuit, the set of performance objectives $\mathcal{S}$ is determined through LLM reasoning over the user's system-level natural language query and the corresponding circuit family-specific performance metric schema defined in the knowledge database, 
with the remaining metrics forming the constraint set $\mathcal{U}$. 
Through an agentic function call, HeaRT locates the corresponding database entries of the identified topologies
using their stored WL 
hash indices
and computes an aggregate rank-based Figure of Merit ($\mathrm{FoM}_{\mathrm{rank}}$) as a fast ranking proxy, 
defined as:
\begin{equation}
\min_{\mathcal{T} \in \phi(\mathcal{U})}
\mathrm{FoM_{rank}}(\mathcal{T}) = \sum_{i \in \mathcal{S}} w_i \times r_i(\mathcal{T}),
\label{eq:fom_rank}
\end{equation}

\noindent
where $w_i$ denotes the metric weight and $\phi(\mathcal{U})$ the feasible search region constrained by 
$\mathcal{U}$:
\begin{equation}
\phi(\mathcal{U}) =
\left\{
\mathcal{T}\ \middle|\ 
|r_j(\mathcal{T}) - \bar{r}_j| \le \epsilon,\ 
\forall j \in \mathcal{U}
\right\},
\label{eq:phi_rank}
\end{equation}
\noindent
where $\bar{r}_j$ denotes the prior rank on the $j$-th metric and $\epsilon$ is a small rank tolerance (set to 3 in our experiments).
If a prior rank is unavailable, 
the constraint is ignored, yielding the full search space along that metric dimension.
The top-$k$ topologies that minimize this score represent promising candidates for the given design goal, 
enabling 
objective-driven retrieval for topology selection.
\section{Experimental Results} \label{sec:Results}

\subsection{Experimental Setup}

\begin{figure}[b]
    \centering
    \vspace{-13pt}
    \includegraphics[width=0.85\linewidth]{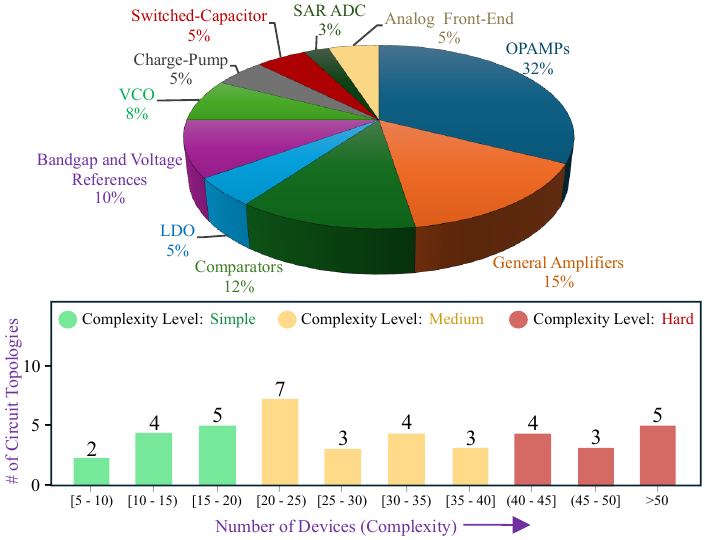}
    \caption{ \small
    Statistical summary of our curated dataset repository.    
    }
    \vspace{-5pt}
    \label{fig:Repository}
    \vspace{-10pt}
\end{figure}

We evaluate HeaRT’s structural circuit decomposition and feedback loop identification on a curated repository of 40 AMS circuits spanning diverse types and complexity levels (Fig.~\ref{fig:Repository}),
provided as flattened SPICE netlists.
Circuits are categorized into three tiers based on transistor count (excluding digital logic components) as a rough proxy for structural complexity: Simple ($<20$ transistors), Medium ($20–40$ transistors), and Hard ($>40$ transistors).
The repository will be open-sourced upon acceptance.
For this task, 
we evaluate HeaRT across a diverse set of LLM backbones spanning
open-source LLMs
(LLaMA-3.3-70B-Instruct ~\cite{grattafiori2024llama,meta2024llama33}, 
DeepSeek-V3.2~\cite{liu2025deepseek}) and proprietary frontier models
(Gemini-2.5 Pro~\cite{gemini2025}, 
GPT-5~\cite{singh2025openai}, and
Claude-4.6-Sonnet~\cite{anthropic2026sonnet46}),
using the F1-based metrics defined in Eq.~\ref{eq:f1_scores}.
To illustrate HeaRT's downstream applicability beyond structural interpretation, 
we further evaluate two system-level circuits: (i) a supply-insensitive relaxation oscillator and (ii) an analog front-end (AFE), 
each under its corresponding design-adaptation scenario (Fig.~\ref{fig:benchmarks}(a)-(b)).
We use Eq.~\eqref{eq:perf_fom} to compute the performance FoM for both scenarios 
and Eq.~\eqref{eq:trs_dvrs} to evaluate design effort reuse for the fixed-architecture case in Scenario 1 (Fig.~\ref{fig:benchmarks}(a)).
Note that since $\mathcal{R}_{\mathrm{effort}}$ measures preservation of prior design effort rather than performance improvement, it is evaluated separately from the FoM. 
All experiments use a 200-simulation budget.
The one-time offline 
tree construction phase employs Claude-4.6-Sonnet,
while downstream interaction through the Interface Agent uses lighter LLM variants (e.g., GPT-4o~\cite{hurst2024gpt}) to reduce LLM inference overhead during online queries.
Schematic simulations are performed using the TSMC180nm process technology with SPICE in Cadence Analog Design Environment (ADE) on a 4-core 3.3 GHz CPU Linux system.

\begin{table}[t]
\centering
\caption{Evaluation of HeaRT across different LLM backbones.}
\vspace{-5pt}
\renewcommand{\arraystretch}{0.75}
\setlength{\tabcolsep}{2.5pt}
\label{tab:heart-sensitivity}
\scriptsize
\begin{tabular}{llcccccc}
\toprule
\multirow{2}{*}[-0.75pt]{\textbf{LLM Backbone}} & \multirow{2}{*}[-0.75pt]{\textbf{Approach}} 
& \multicolumn{3}{c}{$F_{1,\mathrm{subcircuits}}$} 
& \multicolumn{3}{c}{$F_{1,\mathrm{loops}}$} \\[4pt]
& & S & M & H & S & M & H \\
\midrule
\multirow{3}{*}[0.5pt]{\begin{tabular}[c]{@{}l@{}}LLaMA-3.3-\\[1pt]70B-Instruct$^\dag$\end{tabular}}&
HeaRT & \textbf{0.724} & \textbf{0.513} & \textbf{0.336} & \textbf{0.561} & \textbf{0.362} & \textbf{0.191} \\
& Few-shot (Text) & 0.571 & 0.452 & 0.167 & 0.396 & 0.153 & 0.135 \\
& Few-shot (Image) & -- & -- & -- & -- & -- & -- \\
\midrule
\multirow{3}{*}[-1pt]{Gemini-2.5 Pro} & 
  HeaRT & \textbf{0.979} & \textbf{0.952} & \textbf{0.914} & \textbf{0.948} & \textbf{0.891} & \textbf{0.876} \\
& Few-shot (Text) & 0.697 & 0.641 & 0.532 & 0.569 & 0.471 & 0.414 \\
& Few-shot (Image) & 0.681 & 0.612 & 0.469 & 0.591 & 0.554 & 0.397  \\
\midrule
\multirow{3}{*}[-1pt]{DeepSeek-V3.2} & 
  HeaRT & \textbf{0.993} & \textbf{0.974} & \textbf{0.946} & \textbf{0.975} & \textbf{0.917} & \textbf{0.892} \\
& Few-shot (Text) & 0.729 & 0.672 & 0.539 & 0.688 & 0.621 & 0.497 \\
& Few-shot (Image) & 0.718 & 0.637 & 0.493 & 0.702 & 0.626 & 0.464 \\
\midrule
\multirow{3}{*}[-1pt]{GPT-5} & 
  HeaRT & \textbf{0.993} & \textbf{0.973} & \textbf{0.949} & \textbf{0.972} & \textbf{0.931} & \textbf{0.906}  \\
& Few-shot (Text) & 0.722 & 0.669 & 0.578 & 0.691 & 0.638 & 0.523 \\
& Few-shot (Image) & 0.725 & 0.663 & 0.541 & 0.704 & 0.659 & 0.478 \\
\midrule
\multirow{3}{*}[0.5pt]{\begin{tabular}[c]{@{}l@{}}Claude-4.6-\\[1pt]Sonnet\end{tabular}} &
  HeaRT & \textbf{0.997} & \textbf{0.988} & \textbf{0.955} & \textbf{0.984} & \textbf{0.962} & \textbf{0.937} \\
& Few-shot (Text) & 0.762 & 0.709 & 0.684 & 0.741 & 0.675 & 0.614 \\
& Few-shot (Image) & 0.759 & 0.693 & 0.657 & 0.744 & 0.698 & 0.581 \\

\bottomrule
\end{tabular}
\\
\vspace{1.5pt}

\scriptsize
\vspace{1pt}
\textit{
S, M, and H denote Simple ($<$20), Medium (20–40), and Hard ($>$40) circuit complexity tiers in our benchmark repository, measured by transistor count excluding digital logic.
}
\textit{“Text” denotes SPICE netlist input, while “Image” denotes schematic image input.}
$^\dagger$ Vision capability not supported for this model.
\vspace{-12pt}
\end{table}

\subsection{Evaluation of Structural Decomposition and Loop Identification}

Table~\ref{tab:llm_comparison} summarizes the capabilities and limitations of existing LLM-based AMS design frameworks.
We evaluate HeaRT’s ability to perform structural decomposition and feedback loop identification on flattened netlists,
assessing how effectively it recovers circuit structure from raw connectivity,
thereby enabling the hierarchical circuit reasoning paradigm described in
Section~\ref{subsec:need_decomp}.
Using the metrics in Eq.~\ref{eq:f1_scores},
the predicted subcircuits and feedback loops are manually verified against expert-provided ground-truth annotations from experienced analog circuit designers
for the circuits in the benchmark repository (Fig.~\ref{fig:Repository}).
The benchmark repository, including the expert-annotated ground-truth labels used for evaluation, will be publicly released upon acceptance to facilitate reproducibility and further research.
Table~\ref{tab:heart-sensitivity}
reports F1 scores averaged over 10 independent trials
across multiple LLM backbones,
comparing HeaRT against two CoT-based few-shot prompting baselines using netlists and schematic images
as inputs, respectively. 
As shown, performance degrades from Simple to Hard circuits across both tasks as structural complexity increases. 
While schematic image-based reasoning yields marginal gains over netlists on simpler circuits, 
particularly for loop identification where visual feedback paths are clearer, 
schematic images introduce connectivity ambiguity from wire crossings and overlapping nets on larger circuits, 
degrading loop and subcircuit recovery.
Across all circuit complexity tiers, HeaRT consistently improves the F1 scores relative to the baselines,
achieving $\geq13.5\%$ improvement in
$F_{1,\mathrm{subcircuits}}$  and $\geq37.8\%$ in $F_{1,\mathrm{loops}}$,
demonstrating its ability to infer meaningful structure from otherwise unstructured netlist representations.
These gains stem from the combination of our graph-guided partitioning heuristics and LLM reasoning.
As expected, models with stronger reasoning and instruction-following capabilities yield improved performance,
with the best results obtained using Claude-4.6-Sonnet.

\subsection{Integration within Optimization Loops}

The experimental evaluation comprises two incremental design-adaptation scenarios. 
Scenario 1 investigates query-conditioned priority-guided sizing optimization.
We integrate HeaRT with the TuRBO optimizer~\cite{eriksson2019scalable},
which has been widely adopted for analog sizing~\cite{touloupas2021local},
and benchmark it against standalone TuRBO and the LLM-based sizing framework LEDRO~\cite{kochar2025ledro}.
Scenario 2 then evaluates HeaRT’s optimizer‑agnostic and context‑aware decision-making capability by integrating it with several established optimization algorithms commonly used for analog sizing, namely
Differential Evolution (DE)~\cite{vicsan2022automated}, 
the RL-inspired DNN-Opt~\cite{DAC21:Budak:DNN_Opt}, 
and the multi-agent RL-based MA-RL~\cite{zhang2023automated,bao2024multiagent},
alongside TuRBO.
These baselines were chosen to represent different categories of optimization approaches.
In the experiments that follow, we limit HeaRT to a one-time invocation at initialization. 
For a fair convergence comparison under consistent initialization conditions, all optimizers are initialized from the same design point. 
In incremental design adaptation tasks, a uniform starting point reflects practical workflows where an existing design is refined rather than redesigned from scratch.
A detailed analysis of HeaRT’s token usage across the offline and online phases versus circuit complexity tiers is provided in the Appendix~\ref{sec:token_usage} (Fig.~\ref{fig:token_usage_total}).

\begin{figure*}[t]
    \centering
    \includegraphics[width=\linewidth]{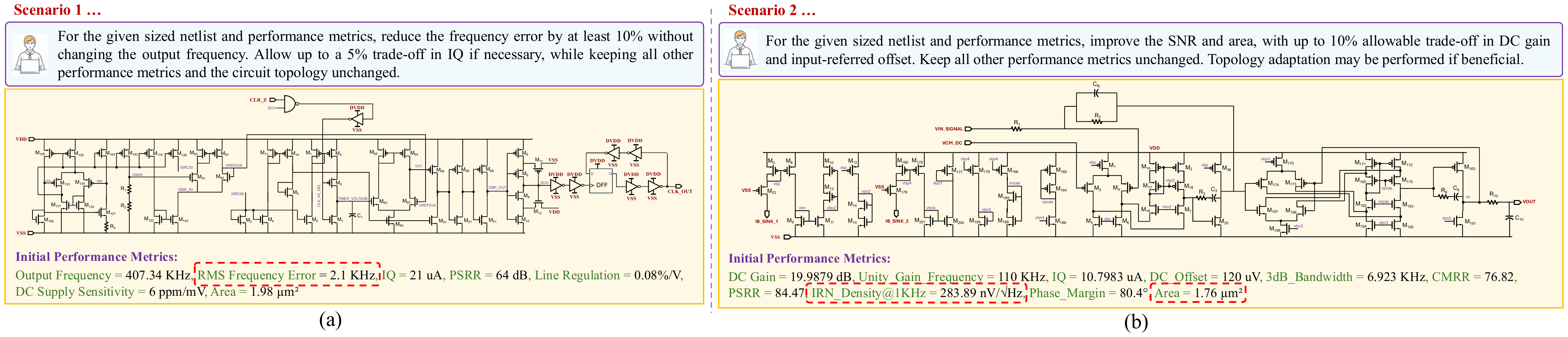}
    \vspace{-17pt}
    \caption{Schematics of (a) Relaxation oscillator and (b) Analog front-end circuit with their corresponding design scenarios.}
    \label{fig:benchmarks}
    \vspace{-5pt}
\end{figure*}

\begin{figure}[htbp]
    \centering
    \vspace{-5pt}
    \includegraphics[width=\linewidth]{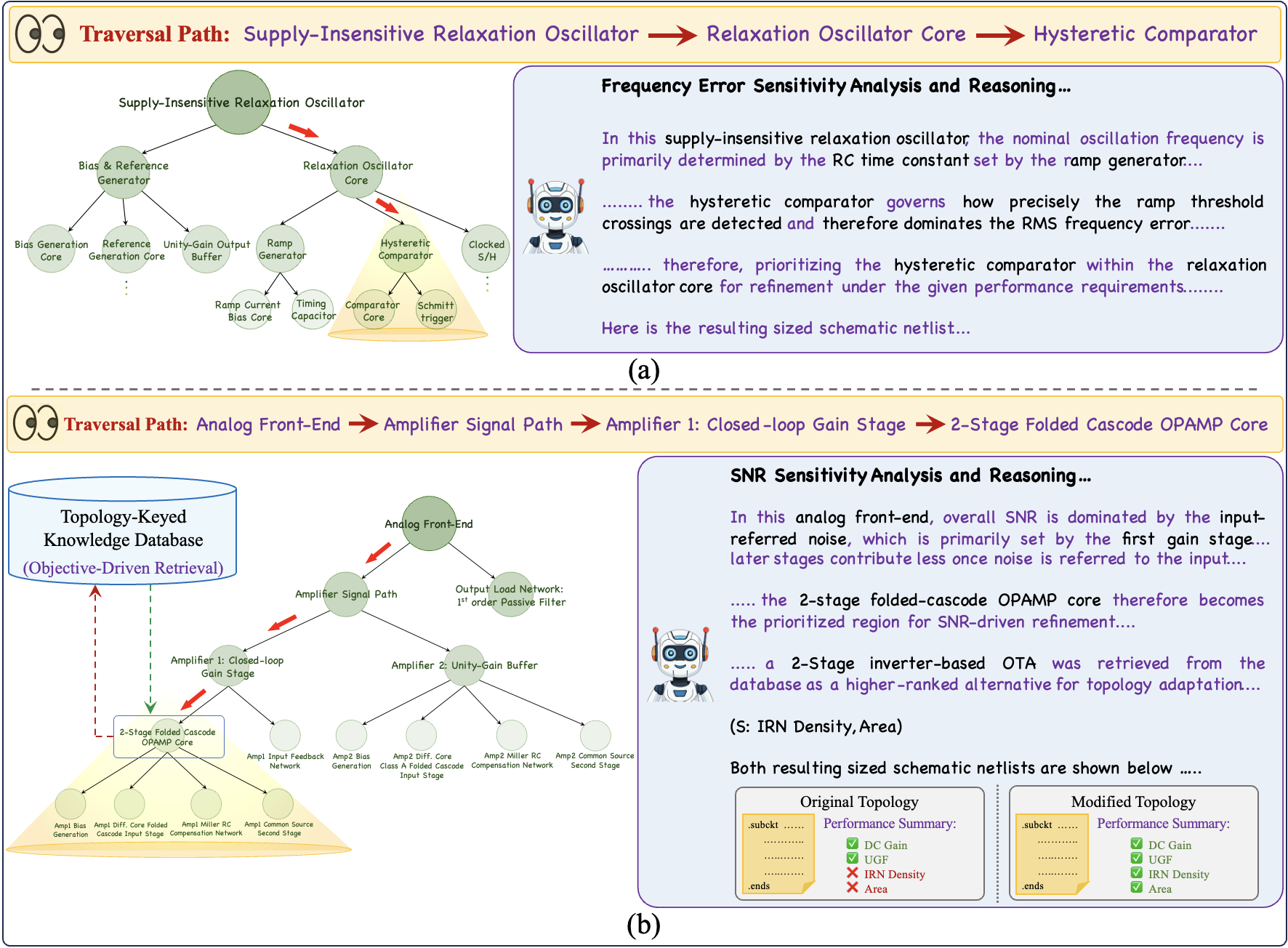}
    \vspace{-15pt}
    \caption{
    Query-conditioned traversal and associated reasoning for (a) Scenario~1 and (b) Scenario~2.
    }    \label{fig:TraversaL_Paths_Reasoning_HeaRT_Tree}
\end{figure}

\subsubsection{Scenario 1: Performance-Shift–Driven Priority-Guided Sizing Optimization}
\begin{figure}[t]
    \centering
    \vspace{5pt}
    \includegraphics[width=0.75\linewidth]{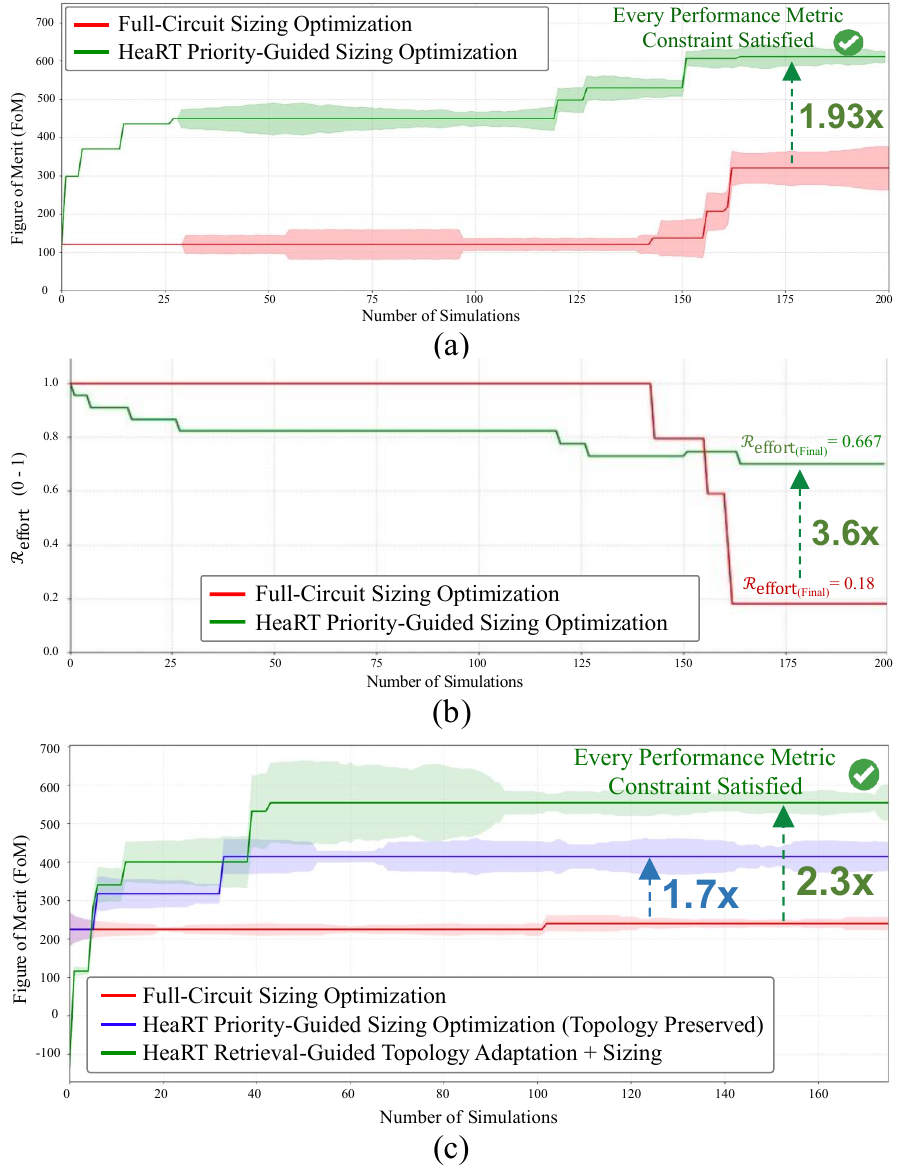}
    \vspace{-3pt}
    \caption{ \small
    Performance profiles for
    (a) FoM and 
    (b) $\mathcal{R}_{\mathrm{effort}}$ in Scenario 1 and 
    (c) FoM in Scenario 2, 
    plotted against simulation count. Shaded regions indicate variation across 10 independent runs.
    }
   \vspace{-2pt} \label{fig:FOM_Reffort_Plots_BO_Scenario_1_and_2}
    \vspace{-10pt}
\end{figure}

As shown in Fig.~\ref{fig:benchmarks}(a), this design adaptation scenario targets a 10\% reduction in frequency error in a supply-insensitive relaxation oscillator while maintaining oscillation frequency, relaxing the $IQ$ tradeoff, and keeping all other key specifications approximately unchanged.
During inference-time query-conditioned traversal on the hierarchical circuit reasoning tree,
HeaRT accurately identifies and prioritizes the 
comparator front-end design variables.
The corresponding traversal path and its reasoning trace are illustrated in Fig.~\ref{fig:TraversaL_Paths_Reasoning_HeaRT_Tree}(a).
The TuRBO optimizer-based sizing procedure is invoked, with HeaRT-derived priority variables $P$ defining the initial bounds: $\mathrm{bounds_{init}}[P] = \textrm{full\ range}$, $\mathrm{bounds_{init}}[\neg P] = \textrm{nominals}$. 
This priority-guided optimization accelerates convergence,
achieving an FoM of $\sim$\textbf{1.93$\times$} and a $\mathcal{R}_{\mathrm{effort}}$ of $\sim\mathbf{3.6\times}$ compared to the standalone TuRBO (Fig.~\ref{fig:FOM_Reffort_Plots_BO_Scenario_1_and_2}(a)-(b)).
Table~\ref{tab:scenario1_results} summarizes the comparative results for Scenario 1 in terms of FoM, convergence speed, 
and real-time token usage.
Unlike approaches that repeatedly invoke LLM reasoning within the optimization loop, HeaRT performs a single real-time invocation to prioritize the search space.

\begin{table}[b]
    \centering
    \vspace{-10pt}
    \caption{Scenario~1: HeaRT+TuRBO vs.\ baselines.}
    \vspace{-5pt}
    \label{tab:scenario1_results}
    \setlength{\tabcolsep}{12pt}
    \renewcommand{\arraystretch}{0.95}
    \scriptsize
    \begin{tabular}{lccc}
        \toprule
        \shortstack{\textbf{Method}\\[0em]} 
        & \shortstack{\textbf{FoM}\\[0em]} 
        & \shortstack{\textbf{\#Sims to} \\ \textbf{Converge}}  
        & \shortstack{\textbf{\#Tokens} \\ \textbf{(Real-Time)}}\\
        \midrule
        TuRBO~\cite{eriksson2019scalable} 
            & 320.71 & 516 & N/A\\
        LEDRO ~\cite{kochar2025ledro} 
            & 395.4 & 354 & 75,212\\
        \textbf{HeaRT+TuRBO} 
            & \textbf{618.97} & \textbf{165}  & \textbf{7,194}\\
        \bottomrule
    \end{tabular}
    \footnotesize
    \\
    \vspace{2pt}
    \textit{GPT-4o used for HeaRT online phase and LEDRO.
    All results averaged over 10 independent trials.}
    \vspace{-12pt}
\end{table}

\begin{table}[t]
\centering
\setlength{\tabcolsep}{8.1pt}
\vspace{-3pt}
\caption{Scenario~2: Impact of HeaRT across different optimizers.}
\vspace{-5pt}
\scriptsize
\begin{tabular}{l|c|c!{\vrule width 1.5pt}c|c}
\hline

\multirow{3}{*}{\textbf{Optimizer}} &
\multirow{3}{*}{\textbf{Eval. Metric}} &
\multicolumn{1}{c!{\vrule width 1.5pt}}{\multirow{3}{*}{\textbf{Baseline}}} &
\multicolumn{2}{c}{\textbf{HeaRT Optimized}}\\
\cline{4-5}
& &  &
\multirow{2}{*}{\textbf{Sizing}} &
\multicolumn{1}{c}{\textbf{Topo+}} \\
& & & 
 &
\multicolumn{1}{c}{\textbf{Sizing}} \\
\hline

\multirow{2}{*}{\textbf{TuRBO~\cite{eriksson2019scalable}}}
  & FoM   & 240.21 & 413.78 & \textbf{555} \\[-1pt]
  & \#Sims to Converge & 401 & \textbf{37} & \textbf{43} \\ 

\hline

\multirow{2}{*}{\textbf{DE~\cite{vicsan2022automated}}}
  & FoM   & 230.21 & 381.82 & \textbf{551} \\[-1pt]
  & \#Sims to Converge & 479 & \textbf{63} & \textbf{68} \\  

\hline

\multirow{2}{*}{\textbf{DNN-Opt~\cite{DAC21:Budak:DNN_Opt}}}
  & FoM   & 269.73 & 467.93 & \textbf{556.6} \\[-1pt]
  & \#Sims to Converge & 286 & \textbf{31} & \textbf{38} \\  

\hline

\multirow{2}{*}{\textbf{MA-RL~\cite{zhang2023automated,bao2024multiagent}}}
  & FoM   & 293.07 & 463.42 & \textbf{556.3} \\[-1pt]
  & \#Sims to Converge & 314 & \textbf{28} & \textbf{39} \\

\hline
\end{tabular}
\vspace{-12pt}
\label{tab:classic_opts}
\end{table}

\begin{table}[t]
\vspace{-2pt}
\caption{Expert–LLM Rank Agreement (Spearman's $\rho_s$ Correlation).}
\vspace{-5pt}
\setlength{\tabcolsep}{4.8pt}
\renewcommand{\arraystretch}{0.9}
\scriptsize
\begin{tabular}{lccccc} 
\toprule 
\shortstack[c]{\textbf{Circuit}\\\textbf{Family}} &
\shortstack[c]{GPT-4o\\[-5pt]\strut} &
\shortstack[c]{Gemini-\\2.5 Pro\strut} &
\shortstack[c]{DeepSeek-\\V3.2\strut} &
\shortstack[c]{GPT-5\\[-5pt]\strut} &
\shortstack[c]{Claude-4.6-\\Sonnet\strut} \\
\midrule
OPAMP & 0.78 / \textbf{0.86} & 0.84 / \textbf{0.90} & 0.87 / \textbf{0.92} & 0.88 / \textbf{0.92} & 0.93 / \textbf{0.95} \\
Comparator & 0.74 / \textbf{0.81} & 0.79 / \textbf{0.88} & 0.89 / \textbf{0.92} & 0.86 / \textbf{0.91} & 0.92 / \textbf{0.93} \\
\bottomrule 
\end{tabular}
\label{tab:retrieval_ranking}
\scriptsize
\centering
\\
\vspace{2pt}
\textit{Results averaged across corresponding family-specific metrics over 10 trials.}
\textit{Each cell reports Base / +Context agreement. Base: few-shot prompting without external grounding; +Context: with knowledge retrieved from ISSCC/JSSC corpus.}
\vspace{-8pt}
\end{table}

\subsubsection{Scenario 2: Knowledge-Based, Performance-Driven Retrieval-Guided Topology Reconfiguration and Sizing}

This scenario targets lower input-referred noise (higher SNR) and 
reduced area under relaxed gain and offset constraints in an analog front-end (Fig.~\ref{fig:benchmarks}(b)).
As shown in Fig.~\ref{fig:TraversaL_Paths_Reasoning_HeaRT_Tree}(b), 
HeaRT identifies the first gain stage as the dominant noise contributor,
with the associated reasoning trace shown alongside.
Its rank-based retrieval, guided by 
$\mathcal{S}$ = \{IRN density, Area\}, 
selects an inverter-based topology as a higher-ranked alternative and generates sized netlists for both designs, 
thereby overcoming architectural bottlenecks inherent in conventional sizing-only workflows.
Fig.~\ref{fig:FOM_Reffort_Plots_BO_Scenario_1_and_2}(c) shows the resulting FoM profiles relative to standalone TuRBO, 
while Table~\ref{tab:classic_opts} summarizes the impact across different baseline optimizers.
As shown, HeaRT achieves an FoM improvement of $\geq 58\%$ through sizing-only optimization, and $\geq 89\%$ through its retrieval-guided topology reconfiguration across all standalone baselines,
while also achieving significantly faster convergence.
Table~\ref{tab:retrieval_ranking} reports the Spearman rank-order correlation coefficient ($\rho_s$) between LLM-derived and human expert rankings 
for 10 representative topologies per family in OPAMPs and comparators, 
comparing Base (few-shot prompting) and +Context (ISSCC/JSSC-grounded) settings to assess the reliability of 
LLM-assigned rankings during topology knowledge database population
(Section~\ref{knowledge_database_management_SP}).
\section{Conclusion} \label{sec:conclusion}

In this paper, we propose HeaRT, a novel hierarchical circuit reasoning tree-based agentic framework for AMS design optimization. 
HeaRT combines top-down graph-guided subcircuit identification and hierarchical abstraction with bottom-up knowledge consolidation, 
enabling structured, context-aware reasoning for AMS design tasks. 
The framework allows natural-language instructions
to steer real-time retrieval while remaining grounded in the hierarchical reasoning structure.
Our experiments validate HeaRT's ability to recover meaningful structure from flattened netlists across a diverse AMS benchmark. 
We further demonstrate its effectiveness in practical design-adaptation scenarios 
involving both topology and sizing optimization,  two critical steps in AMS design.
Future directions include enriching the reasoning tree with simulation metadata and exploring applications such as layout-constraint extraction. 
Overall, HeaRT represents a step toward context-aware, practical AMS design automation.
Upon acceptance, the HeaRT framework and the benchmark repository,
including the expert-annotated ground-truth labels used for evaluation,
will be publicly released as open source.
\newpage

{
\bibliographystyle{unsrt}

\bibliography{ref/Top_sim, ref/reference, ref/ref}
}

\appendices
\section{Token Cost Analysis}\label{sec:token_usage}

\begin{figure}[htbp]
    \centering
    \includegraphics[width=\linewidth]{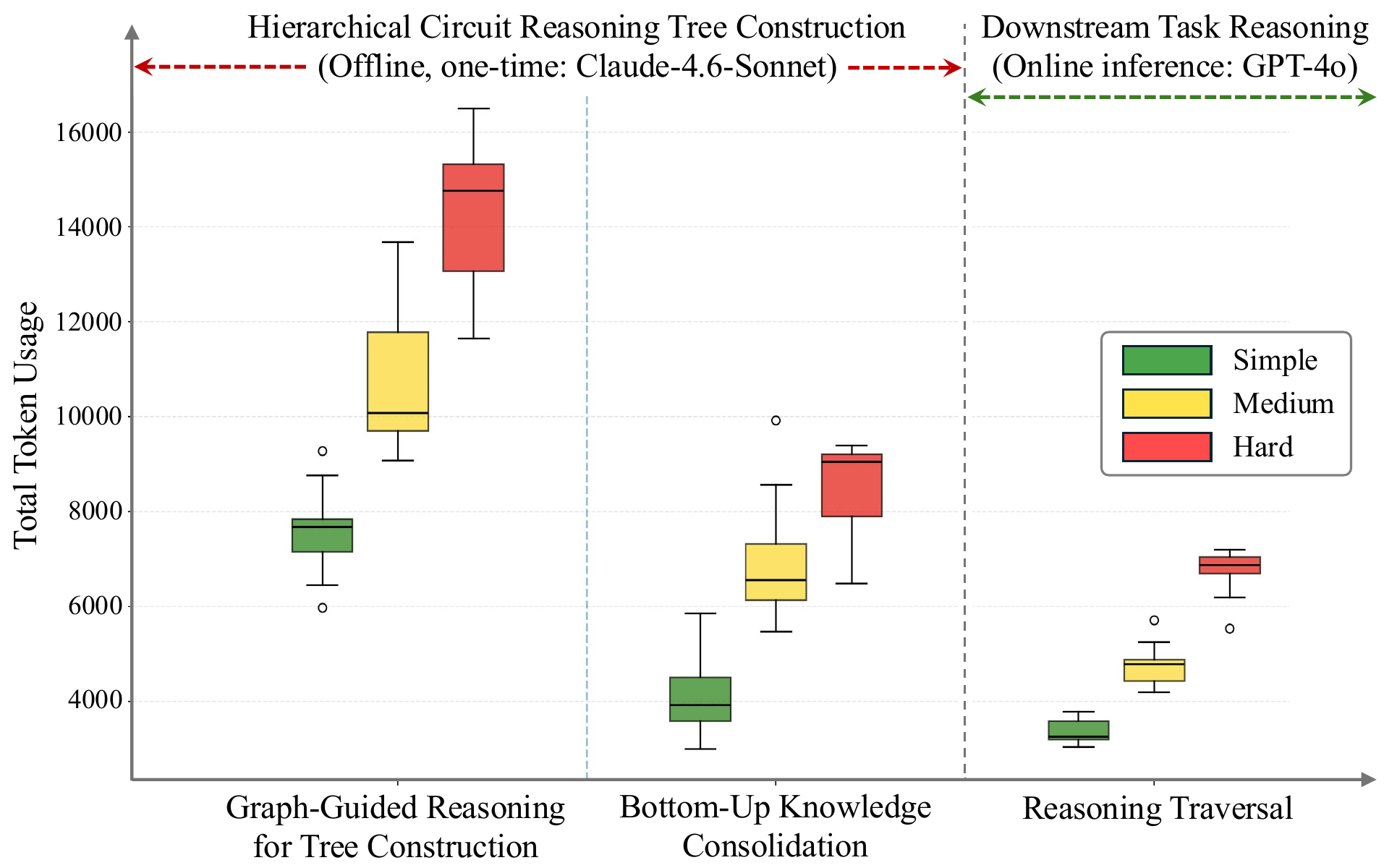}
        \vspace{-10pt}
    \caption{\small Token usage across different stages of the HeaRT workflow.}
    \label{fig:token_usage_total}
    \vspace{-5pt}
\end{figure}

To quantify HeaRT’s overall token cost, we report the total token usage across circuits in our benchmark repository at different stages of the offline and online phases (Fig.~\ref{fig:token_usage_total}), 
grouped by circuit complexity tiers. 
As shown, token usage generally increases with circuit complexity across stages,
with the increase most pronounced in the top-down graph-guided subcircuit decomposition and tree construction stage.
The subsequent bottom-up knowledge consolidation stage shows comparatively smaller variation across complexity tiers.
In contrast, online reasoning stages exhibit noticeably lower variation.
This two-phase design amortizes the higher offline construction cost across downstream applications, 
reducing per-task inference overhead compared to single-shot prompting, particularly for more complex circuits, 
while preserving circuit-level reasoning fidelity.

\noindent
For the experiments reported in Table~\ref{tab:scenario1_results} (Scenario~1) and Table~\ref{tab:classic_opts} (Scenario~2),
the total real-time token usage was approximately 7,194 and 10,216 tokens, respectively.

\section{Additional Implementation Details}

This appendix details the external grounding mechanism for topology knowledge database updates (Fig.~\ref{fig:algorithm_steps}, Section~\ref{knowledge_database_management_SP}). 
We use LlamaIndex~\cite{llamaindex} to index a curated corpus of ISSCC and JSSC publications,
alongside standard analog design textbooks~\cite{razavi2000design,gray2024analysis},
with each document parsed into structured chunks capturing
(i) the document title, (ii) circuit family, (iii) architectural description, and (iv) reported performance characteristics and trade-offs.
Retrieval queries are constructed from the circuit family label and an LLM-generated architectural description of the candidate topology. 
The retrieved context, together with the circuit-family metric schema and current repository state,
is provided as input to the LLM to guide metric-wise ranking updates.
Future extensions could incorporate schematic image-to-graph conversion for richer structural retrieval from analog circuit literature.

Furthermore, while Table~\ref{tab:retrieval_ranking} evaluates the reliability of LLM ranking in a \textit{bulk setting} (all topologies are ranked simultaneously), 
it does not capture the incremental insertion scenario described in Section~\ref{knowledge_database_management_SP},
where a new topology is slotted into an existing repository whose entries already carry metric-wise rankings and concise justification cues reflecting expected performance trade-offs.
To evaluate this, we adopt a leave-one-out protocol over the 10 representative topologies per family used in Table~\ref{tab:retrieval_ranking}: 
in each trial, one topology is held out and inserted into a repository of the remaining 9, with $\rho_s$ measured against human expert ground truth. 
This is repeated across all 10 topologies, with results averaged over 10 independent LLM runs.
Table~\ref{tab:incremental_ranking} reports $\rho_s$ 
under two conditions: 
without and with conditioning on the existing repository state, both grounded in ISSCC/JSSC literature.

\begin{table}[htbp]
\centering
\scriptsize
\caption{\small Incremental Topology Insertion: LLM--Expert Ranking Correlation (Spearman's $\rho_s$, mean $\pm$ std).}
\label{tab:incremental_ranking}
\setlength{\tabcolsep}{5pt}
\renewcommand{\arraystretch}{1.2}
\begin{tabular}{lcccc}
\toprule
\multirow{2}{*}{\textbf{Condition}} & \multicolumn{2}{c}{\textbf{Claude-4.6-Sonnet}} & \multicolumn{2}{c}{\textbf{GPT-5}} \\
\cmidrule(lr){2-3} \cmidrule(lr){4-5}
 & OPAMP & Comparator & OPAMP & Comparator \\
\midrule
\shortstack{w/o Repository \\ Context} & \shortstack{$0.74 \pm 0.09$\\[0pt]}  & \shortstack{$0.70 \pm 0.11$\\[0pt]} & \shortstack{$0.72 \pm 0.10$\\[0pt]} & \shortstack{$0.69 \pm 0.11$\\[0pt]} \\
\shortstack{w/ Repository \\ Context}  & \shortstack{$0.92 \pm 0.04$\\[0pt]}  & \shortstack{$0.90 \pm 0.03$\\[0pt]} & \shortstack{$0.89 \pm 0.06$\\[0pt]} & \shortstack{$0.85 \pm 0.04$\\[0pt]} \\
\bottomrule
\end{tabular}
\\
\vspace{2pt}
\scriptsize
Both conditions are grounded in ISSCC/JSSC literature. Results averaged over 10 independent trials under a leave-one-out protocol over 10 topologies per family.
\end{table}

\section{Efficacy of $\mathrm{FoM_{rank}}$ For Topology Retrieval}
\begin{figure}[htbp]
    \centering
     \vspace{-10pt}
    \includegraphics[width=\linewidth]{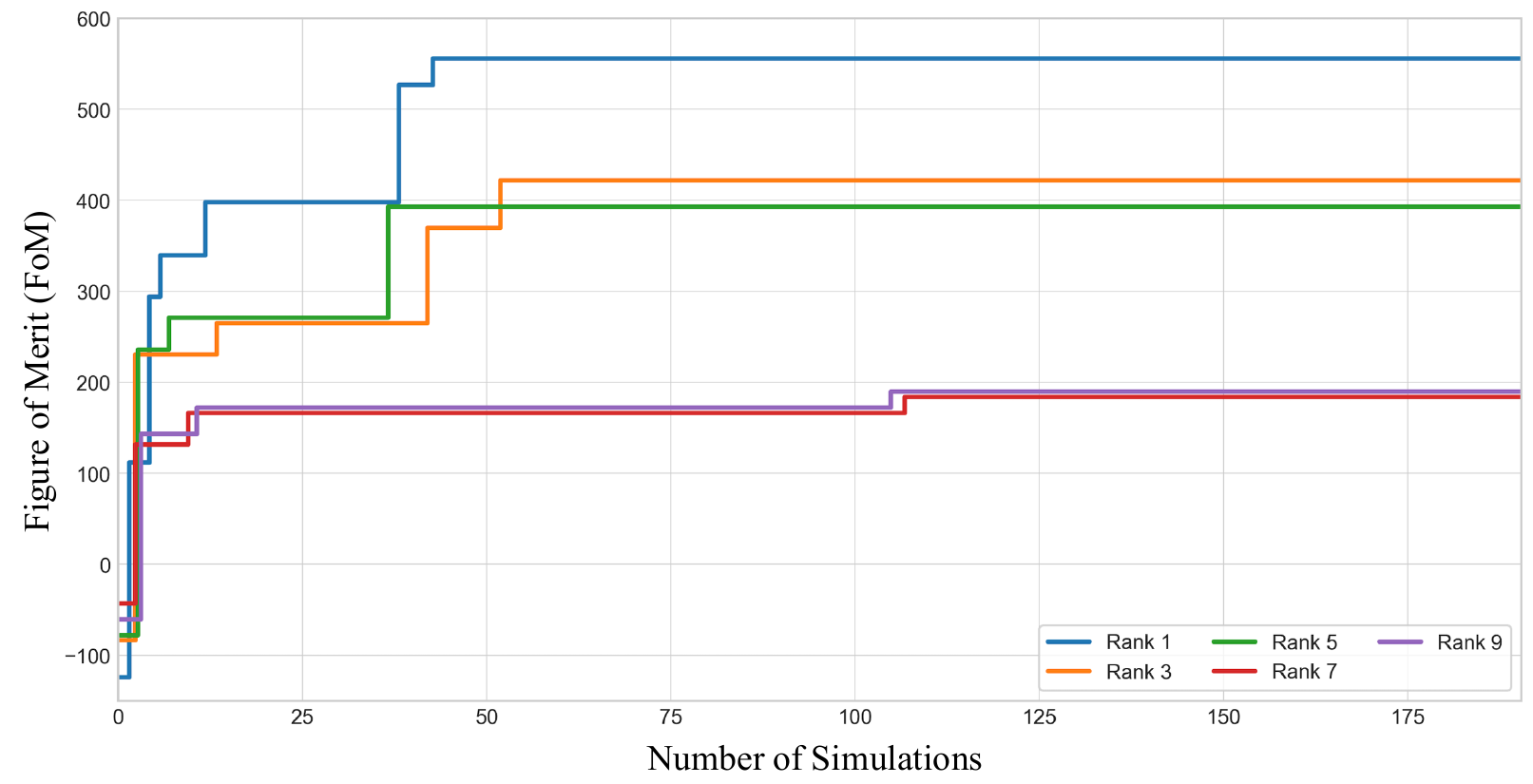}
        \vspace{-10pt}
    \caption{\small Performance FoM Profiles Across Retrieved Topology Ranks.}
    \label{fig:rank_efficacy}
    \vspace{-5pt}
\end{figure}

To further validate the effectiveness of our top-k $\mathrm{FoM_{rank}}$-based topology retrieval strategy, we evaluate the performance profiles across candidates at different rank positions for Scenario 2, under identical optimization settings using TuRBO.
All results are averaged over 10 independent trials.
As shown in Fig.~\ref{fig:rank_efficacy}, higher-ranked candidates generally achieve higher FoM, supporting the utility of our $\mathrm{FoM_{rank}}$-based retrieval as an effective proxy for topology selection.

\end{document}